\documentclass[a4paper,twoside]{article}

\usepackage{epsfig}
\usepackage{subcaption}
\usepackage{calc}
\usepackage{amssymb}
\usepackage{amstext}
\usepackage{amsmath}
\usepackage{amsthm}
\usepackage{multicol}
\usepackage{pslatex}
\usepackage{algorithm2e}
\usepackage[bottom]{footmisc}
\usepackage{apalike}
\usepackage[utf8]{inputenc}
\usepackage[T1]{fontenc}
\usepackage[english]{babel}
\usepackage{microtype}
\usepackage{enumitem}
\usepackage{csquotes}
\usepackage[skip=5pt]{parskip}
\usepackage[hidelinks]{hyperref}
\usepackage{url}
\usepackage{graphicx}
\graphicspath{{images/}}
\usepackage{placeins}
\usepackage{booktabs}
\usepackage{subcaption}
\usepackage{adjustbox}
\usepackage[font=small, labelfont=bf, skip=5pt]{caption}
\usepackage[a4paper, margin=1in]{geometry}
\usepackage{titlesec}
\usepackage{fontawesome}

\titleformat{\section}{\normalfont\Large\bfseries}{\thesection}{1em}{}
\titleformat{\subsection}{\normalfont\large\bfseries}{\thesubsection}{1em}{}
\titleformat{\subsubsection}{\normalfont\normalsize\bfseries}{\thesubsubsection}{1em}{}
\usepackage{lipsum}
\usepackage[capitalize, noabbrev]{cleveref} 
\usepackage{multicol}
\usepackage{multirow}
\usepackage[table]{xcolor}
\newcommand{\good}[1]{\textcolor{blue}{#1}} 
\newcommand{\bad}[1]{\textcolor{red}{#1}}
\newcommand{\iconlink}[2]{%
  \href{#1}{\faGithub\ \textbf{#2}}%
}

\usepackage{SCITEPRESS}
\usepackage{url}

\begin{document}

\setlength{\parindent}{0pt}
\setlength{\parskip}{5pt}

\title{\textbf{GlovEgo-HOI: Bridging the Synthetic-to-Real Gap for Industrial Egocentric Human-Object Interaction Detection}}

\author{
\authorname{
Alfio Spoto\sup{1,2},
Rosario Leonardi\sup{1,2}\orcidAuthor{0009-0001-8693-3826},
Francesco Ragusa\sup{1,2}\orcidAuthor{0000-0002-6368-1910},
and Giovanni Maria Farinella\sup{1,2}\orcidAuthor{0000-0002-6034-0432}
}
\affiliation{\sup{1}Department of Mathematics and Computer Science, University of Catania, Via S. Sofia, 64, 95125 Catania, Italy}
\affiliation{\sup{2}Next Vision s.r.l., Italy}
\email{alfio.spoto@outlook.it, \{rosario.leonardi, francesco.ragusa, giovanni.farinella\}@unict.it}
}


\abstract{Egocentric Human-Object Interaction (EHOI) analysis is crucial for industrial safety, yet the development of robust models is hindered by the scarcity of annotated domain-specific data. We address this challenge by introducing a data generation framework that combines synthetic data with a diffusion-based process to augment real-world images with realistic Personal Protective Equipment (PPE). We present \textit{GlovEgo-HOI}, a new benchmark dataset for industrial EHOI, and \textit{GlovEgo-Net}, a model integrating \textit{Glove-Head} and \textit{Keypoint-Head} modules to leverage hand pose information for enhanced interaction detection.  Extensive experiments demonstrate the effectiveness of the proposed data generation framework and \textit{GlovEgo-Net}.
\\
To foster further research, we release the \textit{GlovEgo-HOI} dataset, augmentation pipeline, and pre-trained models at: 
{\iconlink{https://github.com/NextVisionLab/GlovEgo-HOI}{GitHub project.}}}

\onecolumn \maketitle \normalsize \setcounter{footnote}{0} \vfill

\section{Introduction}
\label{sec:introduction}

Understanding human-object interactions from an egocentric perspective enables intelligent systems to support workers in industrial environments by monitoring their activities, providing contextual information about the objects they interact with, and anticipating potentially dangerous interactions \cite{sener2022assembly101,ragusa2024enigma}.
Unlike generic scenarios where humans interact with common objects, industrial domains are characterized by highly domain-specific objects \cite{ragusa2022meccano,schoonbeek2024industreal}. 
Moreover, specific industrial environments often require the use of personal protective equipment (PPE), such as gloves, which alter the appearance of the hands and make it more difficult to detect and recognize human-object interactions.
Since collecting and manually annotating data in industrial settings is often expensive, time-consuming, and frequently unfeasible due to privacy concerns and proprietary restrictions, data scarcity remains a significant challenge in this field \cite{leonardi2024synthetic}.

\begin{figure*}
    \centering
    \includegraphics[width=\linewidth]{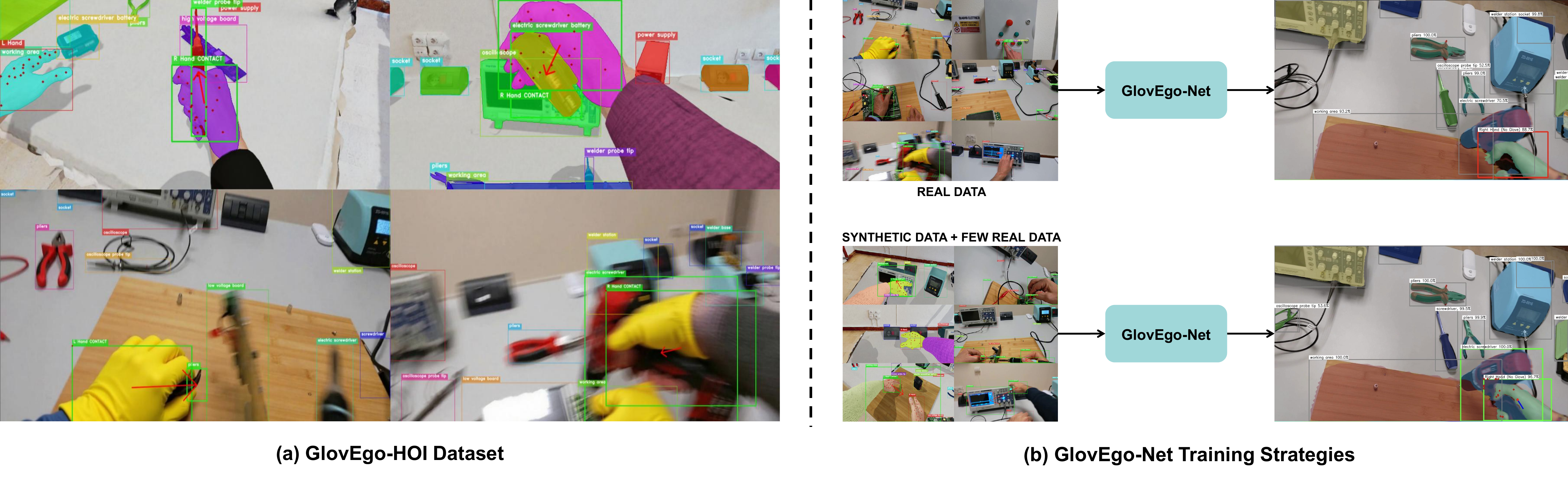} 
    \caption{\textbf{(a) GlovEgo-HOI Dataset}: Examples from our proposed \textbf{GlovEgo-HOI} dataset. The top row shows synthetic images, while the bottom row contains real-world images augmented with PPE via our diffusion-model-based pipeline. \textbf{(b) GlovEgo-Net Training Strategies}: Comparison of \textbf{GlovEgo-Net} performance under different training regimes. (Top) Results with real-world data only (\textit{Real-Only}). (Bottom) Results with our \textit{Synth+Real} approach, using synthetic data for pre-training and real samples for fine-tuning, superior contact state understanding. Our results show that while \textit{Real-Only} models saturate quickly, the inclusion of synthetic keypoints enables \textbf{GlovEgo-Net} to accurately detect interactions even in complex industrial scenes.}
    \label{fig:teaser_figure}
\end{figure*}

To tackle this problem, in this paper, we explore the use of domain-specific synthetic data as a complement to real-world annotated data, enabling the training of models that successfully transfer to real-world scenarios. Although synthetic data can help overcome the lack of real-world annotations, industrial environments present additional layers of complexity. 

To this end, in this work, we introduce a new data generation framework specifically designed for industrial applications. We extend the synthetic data generation pipeline proposed in  \cite{leonardi2022egocentric,leonardi2024exploiting} with automatic annotations of both work gloves and hand keypoints (Figure~\ref{fig:teaser_figure}-a-top). Moreover, to exploit existing annotations of real datasets, we adopted a diffusion model that adds realistic work gloves on existing images, enabling a more accurate representation of industrial scenarios (Figure \ref{fig:teaser_figure}a-bottom). Together, these contributions establish a unified strategy for generating and enriching domain-specific data, enabling robust training of egocentric human-object interaction models. 

Leveraging our proposed data generation framework, we introduce \textit{GlovEgo-HOI}, a new benchmark dataset comprising both synthetic and real-world data for hand-object interaction in industrial contexts (Figure~\ref{fig:teaser_figure}-a).
Building upon the \textit{GlovEgo-HOI} dataset, we also extended the architecture proposed in \cite{leonardi2024exploiting}, with a \textit{Glove-Head} module to detect the presence of personal protective equipment (i.e., work gloves), and a \textit{Keypoint-Head} that integrates fine-grained hand pose information. We refer to this extended model as \textit{GlovEgo-Net}.

Finally, we perform an in-depth analysis of our system, assessing how each design choice, from data enrichment to modular head integration, contributes to improving EHOI detection in industrial domains.

\section{Related Works}
\label{sec:rel_works}

\subsection{Datasets for Egocentric HOI Detection}
Egocentric vision has traditionally relied on kitchen domains with \textit{EPIC-KITCHENS-100} \cite{damen2021rescaling} and \textit{VISOR} \cite{darkhalil2022epic}, which provide rich semantic masks for household tasks. Other foundational works include \textit{EGTEA Gaze+} \cite{li2021eye} for attention-driven actions and \textit{HOI4D} \cite{liu2022hoi4d} for category-level interaction analysis in 4D space. While these datasets advanced the field, they do not reflect the specialized tools and safety constraints of industrial environments. Industrial EHOI detection was pioneered by \textit{MECCANO} \cite{ragusa2021meccano}, which introduced assembly-based interactions. More recently, \textit{ENIGMA-51} \cite{ragusa2024enigma} provided dense temporal annotations of hand-object interactions in industrial laboratory settings. Crucial to our work is \textit{EgoISM-HOI}, an image-based industrial dataset focused on hand-object interactions relying on both real and synthetic data. However, these real-world datasets lack PPE annotations. Our \textit{GlovEgo-HOI} dataset addresses this by augmenting \textit{EgoISM-HOI} with realistic gloves and providing a complementary synthetic subset. Table~\ref{tab:dataset_comparison} compares \textbf{GlovEgo-HOI}with state-of-the-art egocentric datasets for HOI detection in industrial environments.

\begin{table*}[t]
	\centering
	\caption{Comparison of GlovEgo-HOI with state-of-the-art egocentric datasets for HOI detection in industrial environments.}
	\label{tab:dataset_comparison}
	\small 
	\renewcommand{\arraystretch}{1.3}
	\begin{tabular*}{\textwidth}{@{\extracolsep{\fill}}lcccccccc}
		\toprule
		\textbf{Dataset} & \textbf{Year} & \textbf{Type} & \textbf{Settings} & \textbf{Size} & \textbf{Subjects} & \textbf{EHOI} & \textbf{PPE} \\ \midrule
		\textbf{GlovEgo-HOI (Ours)} & \textbf{2026} & \textbf{Synth+Real} & \textbf{Industrial} & \textbf{28,738 imgs} & \textbf{19} & \checkmark & \checkmark \\
		EgoISM-HOI \cite{leonardi2024exploiting} & 2024 & Synth+Real & Industrial & 15,948 imgs & 19 & \checkmark & X \\
		ENIGMA-51 \cite{ragusa2024enigma} & 2024 & Real & Industrial & 22 h & 19 & \checkmark & X \\
		Assembly101 \cite{sener2022assembly101} & 2022 & Real & Industrial-like & 513 h & 53 & X & X \\
        MECCANO \cite{ragusa2021meccano} & 2021 & Real & Industrial-like & 7 h & 20 & \checkmark & X \\  \bottomrule

	\end{tabular*}
\end{table*}

\subsection{Simulators and Synthetic Data}
The high cost of manual annotation has driven interest in simulation platforms like \textit{AI2-THOR} \cite{kolve2017ai2thor}. Research on synthetic hands, such as \textit{SynthHands} \cite{mueller2017realtime} and \textit{ObMan} \cite{hasson2019learning}, has shown that physics-based grasping and randomized backgrounds can effectively train models for real-world tasks. More recently, generative models like \textit{Affordance Diffusion} \cite{ye2023affordance} have explored synthesizing interactions from single images. Our framework extends Unity-based generation to include automatic PPE and pose annotations.

\subsection{Interaction Detection Methods}
EHOI detection typically involves predicting hand side, contact state, and object association \cite{shan2020understanding}. Advanced methods have integrated contact boundaries \cite{zhang2022finegrained} and sequential refinement \cite{fu2022sequential} to improve localization. Multimodal fusion (RGB, depth, and pose) has also proven robust for complex interactions \cite{lu2021egocentric}. Our \textit{GlovEgo-Net} builds on the architecture in \cite{leonardi2024exploiting}, integrating specialized heads for PPE recognition and fine-grained pose information to enhance state classification. 
\section{The GlovEgo-HOI Dataset}
\label{sec:data_gen}

We present \textit{GlovEgo-HOI}, a benchmark dataset for egocentric human-object interaction detection in industrial settings. It comprises a synthetic subset with automatic labeling (\textit{-Synth}) and a real-world subset augmented via a diffusion-based pipeline (\textit{-Real}) to incorporate realistic work gloves. Detailed statistics are provided in Table~\ref{tab:dataset_stats_combined}.

\subsection{GlovEgo-HOI-Synth}
\label{subsec:synth_data}
\begin{figure}[t!]
     \centering
     \includegraphics[width=\columnwidth]{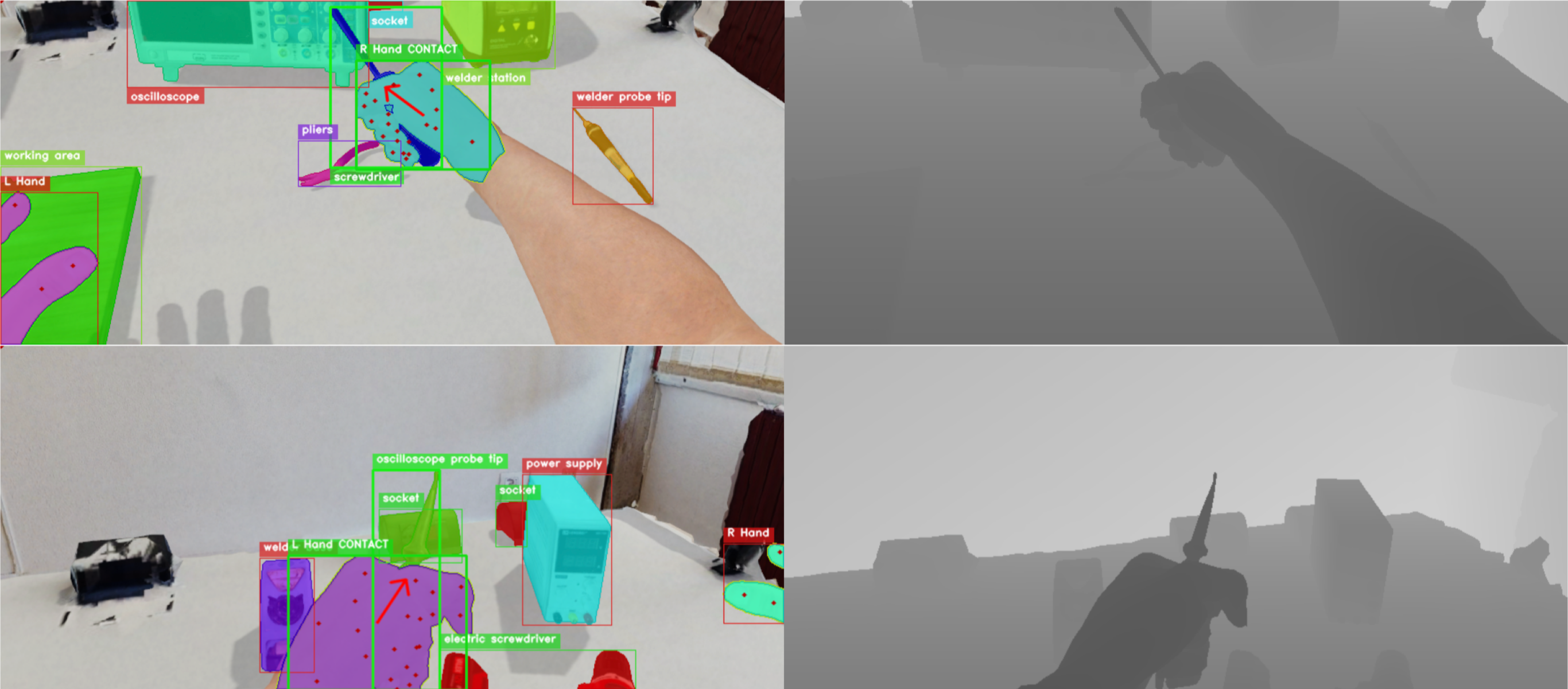}
     \caption{\textit{GlovEgo-HOI-Synth} samples featuring multimodal annotations: RGB, depth, masks, hand keypoints, and glove masks.}
     \label{fig:synth_examples}
\end{figure}

The \textit{-Synth} subset extends the Unity engine and Perception package pipeline from \cite{leonardi2024exploiting}, generating multimodal data including RGB, depth, instance masks, and EHOI annotations. We enhanced this framework with: 
\textbf{(i) Hand Keypoints:} automatic export of 2D coordinates for 21 keypoints per hand instance following the MediaPipe structure \cite{Zhang2020MediaPipeHO}, providing pose data essential for disambiguating complex contact states; 
\textbf{(ii) Glove Annotations:} simulation of PPE by color-shifting 50\% of generated hands to yellow, a technique proven effective for modeling glove-wearing scenarios. Figure~\ref{fig:synth_examples} illustrates samples from this extended pipeline.

\subsection{GlovEgo-HOI-Real}
\label{subsec:real_data}

\begin{figure}[t!]
    \centering
    \includegraphics[width=1\linewidth]{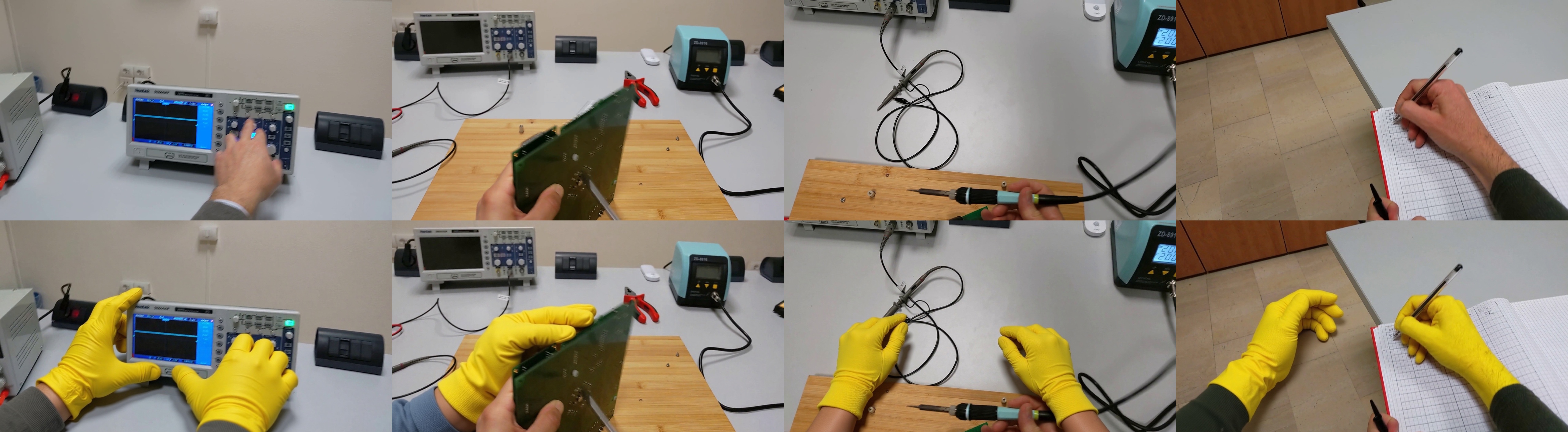}
    \caption{Diffusion artifacts (e.g., background hallucinations) identified and removed via our SSIM-based validation pipeline.}
    \label{fig:flux_error_example}
\end{figure}

\begin{figure*}[t!]
    \centering
    \includegraphics[width=0.9\linewidth]{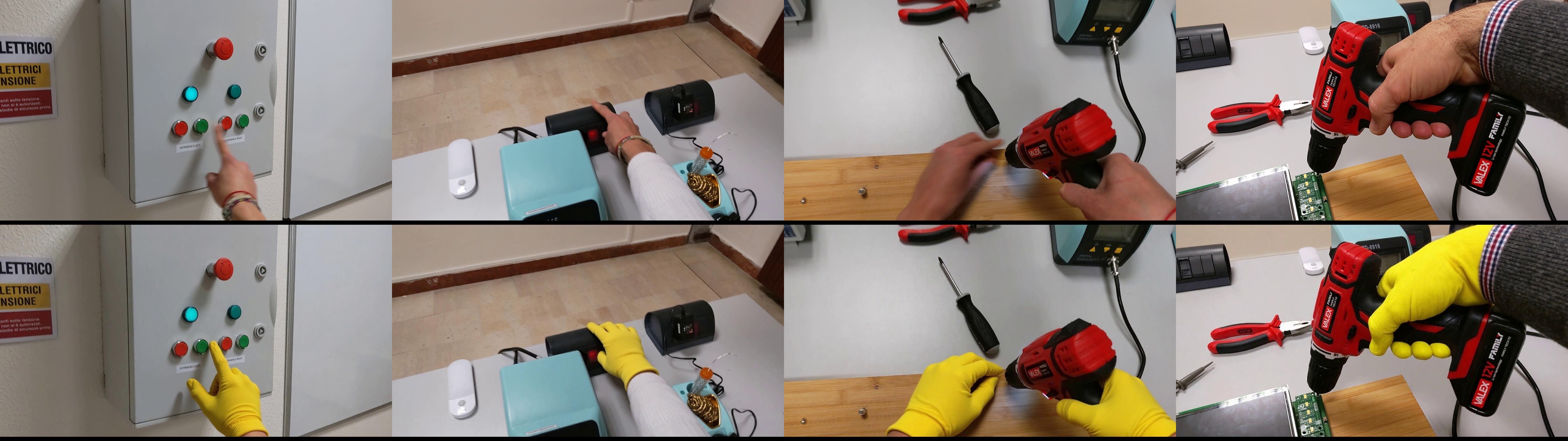}
    \caption{Augmentation pipeline for \textit{GlovEgo-HOI-Real}: original frames (top) and corresponding FLUX-augmented outputs (bottom).}
    \label{fig:flux_examples}
\end{figure*}

For \textit{GlovEgo-HOI-Real}, we utilize 15,948 manually annotated frames from \textit{EgoISM-HOI} \cite{leonardi2024exploiting}, featuring 19 subjects in industrial procedures. To integrate PPE context, we augmented these frames via \textbf{FLUX.1-Kontext-dev} \cite{flux2023} using the prompt: \textit{"Add a yellow working glove(s) to the hand(s) in the image, while preserving the original number of hands and their exact positions."}

To ensure augmentation reliability and mitigate artifacts (e.g., semantic hallucinations or background inconsistencies shown in Fig.~\ref{fig:flux_error_example}), we implemented an automated validation based on background consistency. For each original-augmented pair, we mask the hand regions and calculate the Structural Similarity Index (SSIM) \cite{wang2004image} between background-only images. Pairs with $SSIM < 0.95$ are discarded. This empirical threshold ensures structural integrity and scene consistency, forcing the model to focus exclusively on added PPE features. The resulting filtered set forms \textit{GlovEgo-HOI-Real} (Fig.~\ref{fig:flux_examples}).

\begin{table*}[t!]
	\centering
	\caption{Full statistics of the \textbf{GlovEgo-HOI} benchmark (Synthetic and Real subsets).}
	\label{tab:dataset_stats_combined}
	\begin{adjustbox}{max width=\textwidth}
		\begin{tabular}{lrrrrrrr} 
			\toprule
			\textbf{Split}         & \textbf{\#images} & \textbf{\#hands} & \textbf{\#EHOIs} & \textbf{\#left hands} & \textbf{\#right hands} & \textbf{\#objects} & \textbf{\%glove hands} \\
			\midrule
			Train (Synth)          & 8,953             & 14,191           & 7,240            & 7,209                 & 6,982                  & 50,050             & 50.62                \\
			Val (Synth)            & 2,558             & 4,112            & 2,099            & 2,083                 & 2,029                  & 14,224             & 49.73                \\
			Test (Synth)           & 1,279             & 2,011            & 1,047            & 1,003                 & 1,008                  & 6,982              & 49.38                \\
			\textbf{Total (Synth)} & \textbf{12,790}   & \textbf{20,314}  & \textbf{10,386}  & \textbf{10,295}       & \textbf{10,019}        & \textbf{71,256}    & \textbf{50.32}       \\
			\midrule
			Train (Real)           & 1,010             & 1,686            & 1,262            & 758                   & 928                    & 6,689              & 20.28                 \\
			Val (Real)             & 3,717             & 5,622            & 3,867            & 2,577                 & 3,045                  & 20,916             & 20.26                 \\
			Test (Real)            & 11,221            & 16,850           & 11,403           & 7,743                 & 9,107                  & 62,356             & 16.56                 \\
			\textbf{Total (Real)}  & \textbf{15,948}   & \textbf{24,158}  & \textbf{16,532}  & \textbf{11,078}       & \textbf{13,080}        & \textbf{89,961}    & \textbf{17.68}        \\
			\bottomrule
		\end{tabular}
	\end{adjustbox}
\end{table*}
\section{The GlovEgo-Net Architecture}
\label{sec:proposed_method}

\begin{figure*}[h!]
    \centering
    \includegraphics[width=1\linewidth]{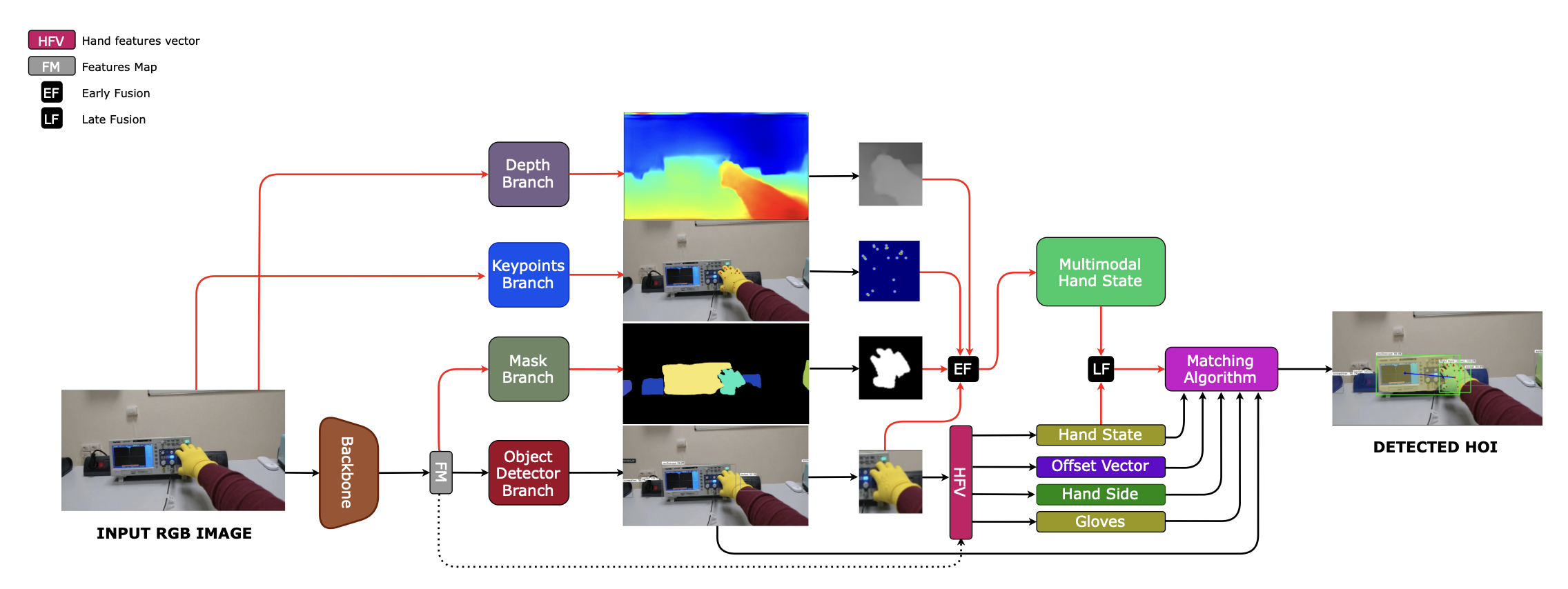}
    \caption{Overview of the \textbf{GlovEgo-Net} architecture. The system processes RGB images to extract multimodal features, fusing pose heatmaps, depth, and masks through Early Fusion (EF) and Late Fusion (LF) to output final EHOI quadruples.}
    \label{fig:architecture}
\end{figure*}

\textbf{GlovEgo-Net} extends the multimodal architecture in \cite{leonardi2024exploiting} by integrating a \textit{Glove-Head} for PPE detection and a \textit{Keypoint-Head} for pose information (Fig.~\ref{fig:architecture}). The system uses a ResNet-101 \cite{he2016deep} backbone with a Feature Pyramid Network (FPN) \cite{lin2017feature} to extract multi-scale feature maps for parallel branches: Faster R-CNN \cite{ren2015faster} for detection, Mask R-CNN \cite{he2017mask} for segmentation, and a monocular depth estimation branch based on MiDaS \cite{ranftl2022toward}. For each hand instance, a 1024-dimensional \textit{Hand Feature Vector} (HFV) is extracted via RoI pooling to feed four MLP-based attribute heads: \textbf{Side} (left/right), \textbf{State} (appearance-based contact), \textbf{Offset Vector} ($\langle v_x, v_y, m \rangle$ for association), and the new \textbf{Glove Head} for PPE status.

\subsection{Hand Keypoints for Contact Estimation}
\label{ssubsec:keypoint_fusion}
We integrate a keypoint branch \cite{he2017mask} predicting $N=21$ coordinates per hand. Pose heatmaps provide fine-grained geometric cues that disambiguate contact states where visual appearance is insufficient. These heatmaps are concatenated with the hand's RGB crop, segmentation mask, and depth map in an \textbf{Early Fusion} (EF) block. This four-modality tensor is processed by an EfficientNetV2-based \cite{tan2021efficientnetv2} \textit{Multimodal Hand State classifier} for robust interaction detection.

\subsection{Final Egocentric Human-Object Interaction Prediction}
\label{subsec:matching_loss}
The final EHOI prediction is obtained through a two-stage process. First, the final contact state is obtained via \textit{Late Fusion}, which combines two sources of information: the Hand State prediction, relying exclusively on the hand's appearance features, and the Multimodal Hand State prediction, which leverages additional modalities such as depth and hand keypoints. This fusion strategy yields a more stable and accurate contact estimation than either modality alone.

In the second stage, a custom \textit{Matching Algorithm} maps hands predicted as in contact with objects in the scene. It takes all model outputs, including detected hands and objects, and hand attributes (Side, Glove, final Contact State, and the Offset Vector $\langle v_x, v_y, m \rangle$). For each hand classified as \textit{in contact}, the Offset Vector is used to project an interaction point, and the object whose bounding box center is closest to this point is selected as the corresponding active target. The system outputs EHOIs as quadruples $\langle \text{hand, contact state, active object, glove} \rangle$, with glove denoting the predicted glove status for the hand.

The \textbf{GlovEgo-Net} architecture is trained end-to-end. The total loss $L_{\text{total}}$ is defined as the sum of all component losses:
\begin{equation}
\label{eq:total_loss}
\begin{aligned}
L_{\text{total}} ={}& L_{\text{backbone}} + L_{\text{depth}} + L_{\text{side}} + L_{\text{contact}} \\
                   & + L_{\text{offset}} + L_{\text{kpt}} + L_{\text{glove}}
\end{aligned}
\end{equation}
where $L_{\text{backbone}}$ includes the standard R-CNN losses; $L_{\text{contact}}$ is the composite contact loss defined in \cite{leonardi2024exploiting}; and $L_{\text{kpt}}$ and $L_{\text{glove}}$ are the binary cross-entropy losses for the new keypoint and glove prediction modules.
\section{Experiments}
\label{sec:experiments}

In this section, we present the experimental analysis conducted to assess the effectiveness of the proposed \textbf{GlovEgo-Net}. Our experiments are designed to answer two questions: (1) how leveraging synthetic data from our \textbf{GlovEgo-HOI} during training influences model performance compared to training exclusively on real data, and (2) how the integration of our newly \textit{Keypoint Head} and \textit{Glove Head} modules affects the overall model performance.

\subsection{Experimental Setup}
\label{subsec:exp_setup}

\noindent\textbf{Datasets} We conduct all experiments using our \textbf{GlovEgo-HOI} dataset, which is split into \textit{GlovEgo-HOI-Synth} (synthetic data) and \textit{GlovEgo-HOI-Real} (real-world data). Detailed statistics for these splits are presented in Section~\ref{sec:data_gen}.

\noindent\textbf{Training Strategies} We compare three different training strategies to assess the benefits of using synthetic data for improving model performance:
\begin{itemize}
    \item \textbf{Synth-Only}: models trained exclusively on \textit{GlovEgo-HOI-Synth} and evaluated on the \textit{GlovEgo-HOI-Real} test set;
    \item \textbf{Real-Only}: models trained exclusively on \textit{GlovEgo-HOI-Real} training set and evaluated on the \textit{GlovEgo-HOI-Real} test set;
    \item \textbf{Synth+Real}: models pre-trained on \textit{GlovEgo-HOI-Synth} and then fine-tuned on \textit{GlovEgo-HOI-Real}.
\end{itemize}
 
\noindent\textbf{Metrics} 
We evaluate model performance using different metrics introduced in \cite{shan2020understanding} and \cite{leonardi2024exploiting}, all derived from standard \textit{Average Precision} (AP) and \textit{mean Average Precision} (mAP), computed at an \textit{IoU} threshold of $0.5$. This evaluation protocol measures distinct aspects of the model’s EHOI predictions, which we further extend to include our new industrial safety task. Specifically, we report:

\begin{itemize}
    \item \textbf{AP Hand:} Standard \textit{Average Precision} for hand detection.
    \item \textbf{AP Hand+Side:} \textit{Average Precision} for hand detection, counting a detection as correct only if the hand side (left/right) is also correctly predicted.
    \item \textbf{AP Hand+State:} \textit{Average Precision} for hand detection, counting a detection as correct only if the contact state (contact/no-contact) is also correctly predicted.
    \item \textbf{AP Hand+Glove:} \textit{Average Precision} for hand detection, counting a detection as correct only if the PPE status (glove/no glove) is correctly predicted.
    \item \textbf{mAP Hand+Obj:} \textit{mean Average Precision} for the detection of $\langle \text{hand, active object} \rangle$ pairs, where a detection is considered correct only if both the hand and the associated object are correctly localized and the object is correctly classified.
    \item \textbf{mAP Hand+All:} \textit{mean Average Precision} combining all previous metrics, where a detection is considered correct only if all corresponding elements (side, state, object, and PPE) are correctly predicted.
\end{itemize}

\noindent\textbf{Inference Latency Analysis}
To evaluate the industrial feasibility of \textit{GlovEgo-Net}, we measured its inference latency on an NVIDIA Tesla V100S GPU (32GB). The model achieves an average processing time of \textbf{148.23~ms} per frame, resulting in a throughput of \textbf{6.75~FPS}.

\begin{table*}[t!]
	\centering
	\caption{Performance on the \textit{GlovEgo-HOI-Real} test set comparing three training strategies: \textit{Real-Only}, \textit{Synth-Only}, and \textit{Synth+Real}. Improvement indicates the gain (\good{in blue}) or loss (\bad{in red}) of \textit{Synth+Real} over \textit{Real-Only}.}
	\label{tab:real_data_percentage_ablation}
	\begin{adjustbox}{max width=\textwidth}
		\begin{tabular}{llcccccc}
			\toprule
			\textbf{Pre-train Synth} & \textbf{Real \%}  & \textbf{AP Hand} & \textbf{AP Hand+Side} & \textbf{AP Hand+Glove} & \textbf{AP Hand+State} & \textbf{mAP Hand+Obj} & \textbf{mAP Hand+All} \\
			\midrule
			Yes (Synth-Only)           & 0\%                    & 82.96            & 61.76               & 72.82                & 30.25                & 8.41                & 6.22                \\
			\midrule
			No (Real-Only)              & \multirow{2}{*}{10\%}  & \textbf{92.44}   & \textbf{88.01}      & 78.38                & 43.43                & 12.78               & 12.07               \\
			Yes (Synth+Real)            &                        & 91.11            & 85.19               & \textbf{79.72}       & \textbf{46.72}       & \textbf{13.60}      & \textbf{12.34}      \\
			\textit{Improvement}                 &                        & \textcolor{red}{-1.33}      & \textcolor{red}{-2.82}         & \textcolor{blue}{+1.34}         & \textcolor{blue}{+3.29}         & \textcolor{blue}{+0.82}         & \textcolor{blue}{+0.27}         \\
			\midrule
			No (Real-Only)              & \multirow{2}{*}{25\%}  & \textbf{92.44}   & 88.01               & 78.38                & 43.16                & 12.73               & 12.00               \\
			Yes (Synth+Real)            &                        & 92.23            & \textbf{88.86}      & \textbf{82.06}       & \textbf{48.47}       & \textbf{13.56}      & \textbf{13.10}      \\
			\textit{Improvement}                 &                        & \textcolor{red}{-0.21}      & \textcolor{blue}{+0.85}         & \textcolor{blue}{+3.68}         & \textcolor{blue}{+5.31}         & \textcolor{blue}{+0.83}         & \textcolor{blue}{+1.10}         \\
			\midrule
			No (Real-Only)              & \multirow{2}{*}{50\%}  & 92.44            & 88.01               & 78.38                & 42.85                & 12.50               & 11.84               \\
			Yes (Synth+Real)            &                        & \textbf{94.22}   & \textbf{90.50}      & \textbf{84.07}       & \textbf{48.97}       & \textbf{15.50}      & \textbf{14.45}      \\
			\textit{Improvement}                 &                        & \textcolor{blue}{+1.78}     & \textcolor{blue}{+2.49}         & \textcolor{blue}{+5.69}         & \textcolor{blue}{+6.12}         & \textcolor{blue}{+3.00}         & \textcolor{blue}{+2.61}         \\
			\midrule
			No (Real-Only)              & \multirow{2}{*}{100\%} & \textbf{95.90}   & \textbf{93.18}      & \textbf{84.92}       & 49.42                & 19.04               & 18.12               \\
			Yes (Synth+Real)            &                        & 95.05            & 91.93               & 84.84                & \textbf{51.16}       & \textbf{19.77}      & \textbf{19.06}      \\
			\textit{Improvement}                 &                        & \textcolor{red}{-0.85}      & \textcolor{red}{-1.25}         & \textcolor{red}{-0.08}          & \textcolor{blue}{+1.74}         & \textcolor{blue}{+0.73}         & \textcolor{blue}{+0.94}         \\
			\bottomrule
		\end{tabular}
	\end{adjustbox}
\end{table*}

\subsection{Impact of Synthetic Data on Model Performance}
\label{subsec:data_efficiency}

We evaluate the impact of synthetic data by comparing \textit{Real-Only}, \textit{Synth-Only}, and \textit{Synth+Real} strategies (Table~\ref{tab:real_data_percentage_ablation}). While \textit{Synth-Only} training provides a reasonable initialization for detection ($83.87\%$ AP Hand), low interaction scores (e.g., $6.22\%$ mAP Hand+All) confirm that domain adaptation is essential to bridge the sim-to-real gap.

Our results highlight a critical "performance plateau" in purely real-world training; increasing \textit{Real-Only} data from 10\% to 50\% fails to significantly improve contact estimation, which remains capped around $43\%$. Conversely, the \textit{Synth+Real} strategy consistently overcomes this saturation. At the 50\% split, our approach achieves a gain of \textbf{+6.12\%} in \textit{AP Hand+State} and \textbf{+2.61\%} in \textit{mAP Hand+All}. 

Even with 100\% of real data, the \textit{Synth+Real} model reaches the highest overall performance (\textbf{19.06\%} mAP), surpassing the \textit{Real-Only} baseline ($18.12\%$). This confirms that geometric information from synthetic keypoints and depth maps provides an upper bound that appearance-based training alone cannot achieve.

\subsection{Ablation Study: Impact of Keypoint Head}
\label{subsec:ablation_study}
Table~\ref{tab:ablation_study_synth_real} evaluates the contribution of the \textit{Keypoint Head} by comparing the full \textit{GlovEgo-Net} against a variant lacking pose information. On synthetic data, the full model consistently improves performance, particularly in complex interaction metrics where \textit{mAP Hand+Obj} increases from $60.28\%$ to $\textbf{62.95\%}$. While pose information introduces minor overhead in raw localization, it provides a superior signal for structured interaction, resulting in a $\textbf{+2.67\%}$ gain in \textit{mAP Hand+All}.

On real data, the Keypoint Head maintains its effectiveness even in zero-shot transfer, boosting \textit{mAP Hand+Obj} from $6.61\%$ to $\textbf{8.41\%}$. Although the base model shows lower sensitivity to geometric domain shifts in individual attributes (Side and State), the integration of hand keypoints proves essential for accurately linking interactions to the correct objects in complex industrial scenes.

\begin{table}[htbp] %
    \centering
    \footnotesize 
    \setlength{\tabcolsep}{2.5pt} 
    \caption{Ablation study on the \textit{Keypoint Head}. Models are trained on \textit{Synth} and evaluated on both domains (Zero-shot on Real).}
    \label{tab:ablation_study_synth_real}
    \begin{tabular}{lcc|cc}
        \toprule
        \multirow{2}{*}{\textbf{Metric}} & \multicolumn{2}{c}{\textbf{Tested on Synth}} & \multicolumn{2}{c}{\textbf{Tested on Real}} \\
        \cmidrule(lr){2-3} \cmidrule(lr){4-5}
                       & \begin{tabular}[c]{@{}c@{}}w/o\\ Kpts\end{tabular} & \begin{tabular}[c]{@{}c@{}}Full\\ Net\end{tabular} & \begin{tabular}[c]{@{}c@{}}w/o\\ Kpts\end{tabular} & \begin{tabular}[c]{@{}c@{}}Full\\ Net\end{tabular} \\
        \midrule
        AP Hand        & \textbf{98.72} & 97.84          & 81.69          & \textbf{82.96} \\
        AP Hand+Side   & \textbf{97.83} & 97.77          & \textbf{66.23} & 61.76          \\
        AP Hand+Glove  & \textbf{98.72} & 97.84          & 68.34          & \textbf{72.82} \\
        AP Hand+State  & 92.09          & \textbf{92.43} & \textbf{31.47} & 30.25          \\
        mAP Hand+Obj   & 60.28          & \textbf{62.95} & 6.61           & \textbf{8.41}  \\
        mAP Hand+All   & 60.18          & \textbf{62.85} & 5.44           & \textbf{6.22}  \\
        \bottomrule
    \end{tabular}
\end{table}

\begin{figure*}[t!]
    \centering
    \includegraphics[width=0.9\linewidth]{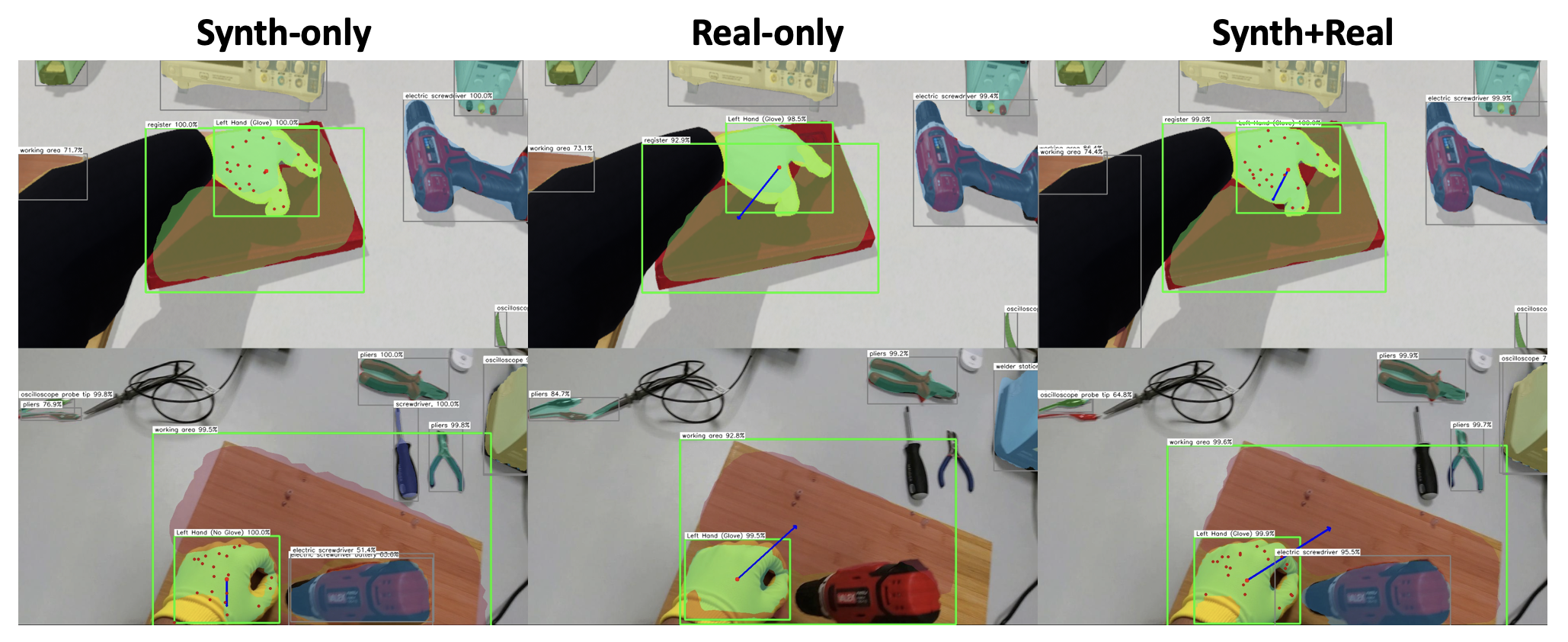}  
    \caption{
        Qualitative comparison of the three training regimes on \textit{GlovEgo-HOI}.
        \textbf{Top row (\textit{GlovEgo-HOI-Synth}):} Predictions from models trained with \textit{Synth-Only}, \textit{Real-Only}, and \textit{Synth+Real}, all evaluated on \textit{GlovEgo-HOI-Synth}.
        \textbf{Bottom row (\textit{GlovEgo-HOI-Real}):} The same three training regimes evaluated on \textit{GlovEgo-HOI-Real}.
    }
    \label{fig:qualitative_overview}
\end{figure*}

\subsection{Qualitative Results}
\label{subsec:qualitative}

\begin{figure*}[t!]
    \centering
    \subcaptionbox{GlovEgo-Net w/o Kpts\label{fig:zeroshot_a}}{%
        \includegraphics[width=0.45\linewidth]{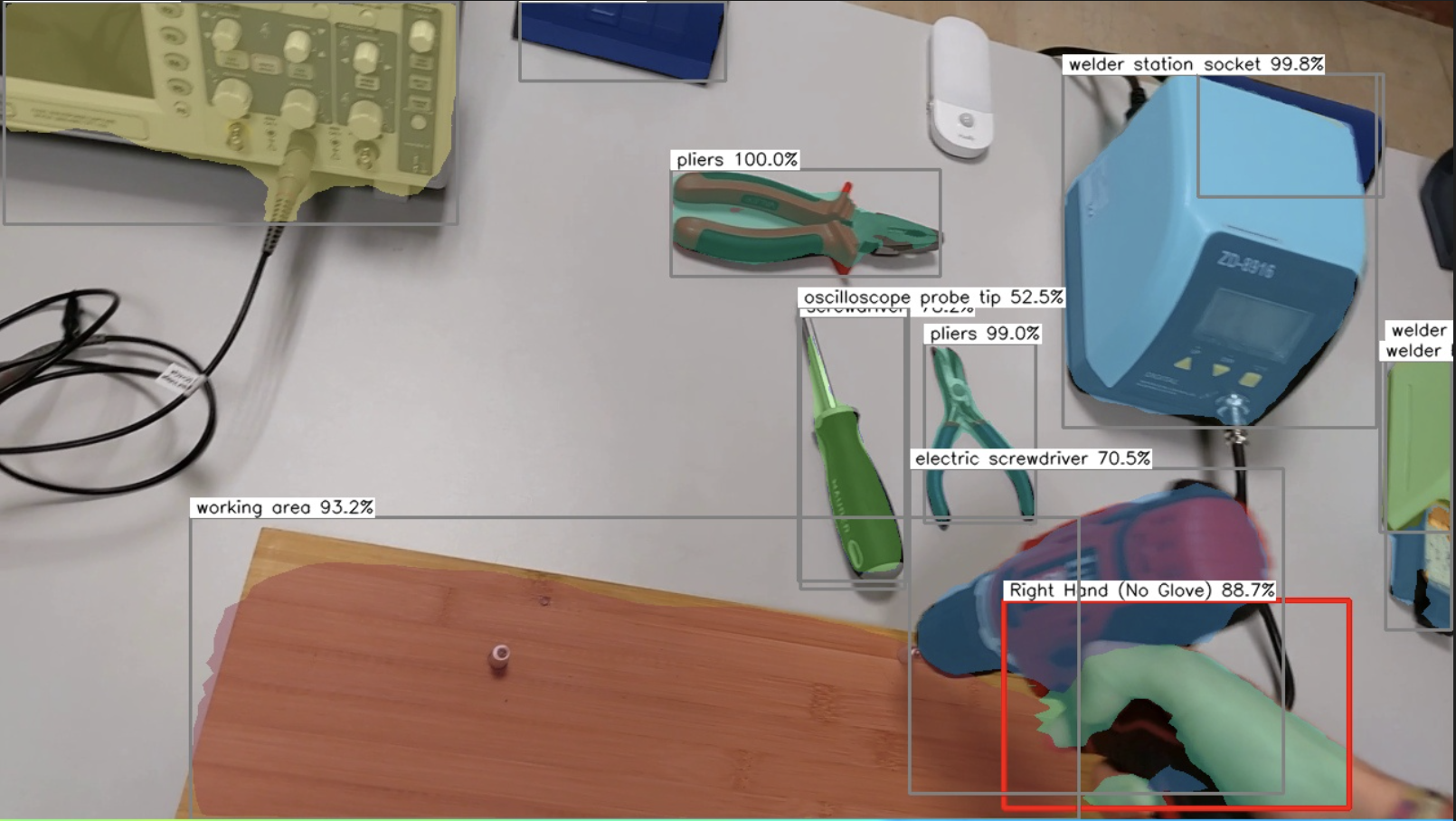} 
    }%
    \subcaptionbox{GlovEgo-Net\label{fig:zeroshot_b}}{%
        \includegraphics[width=0.45\linewidth]{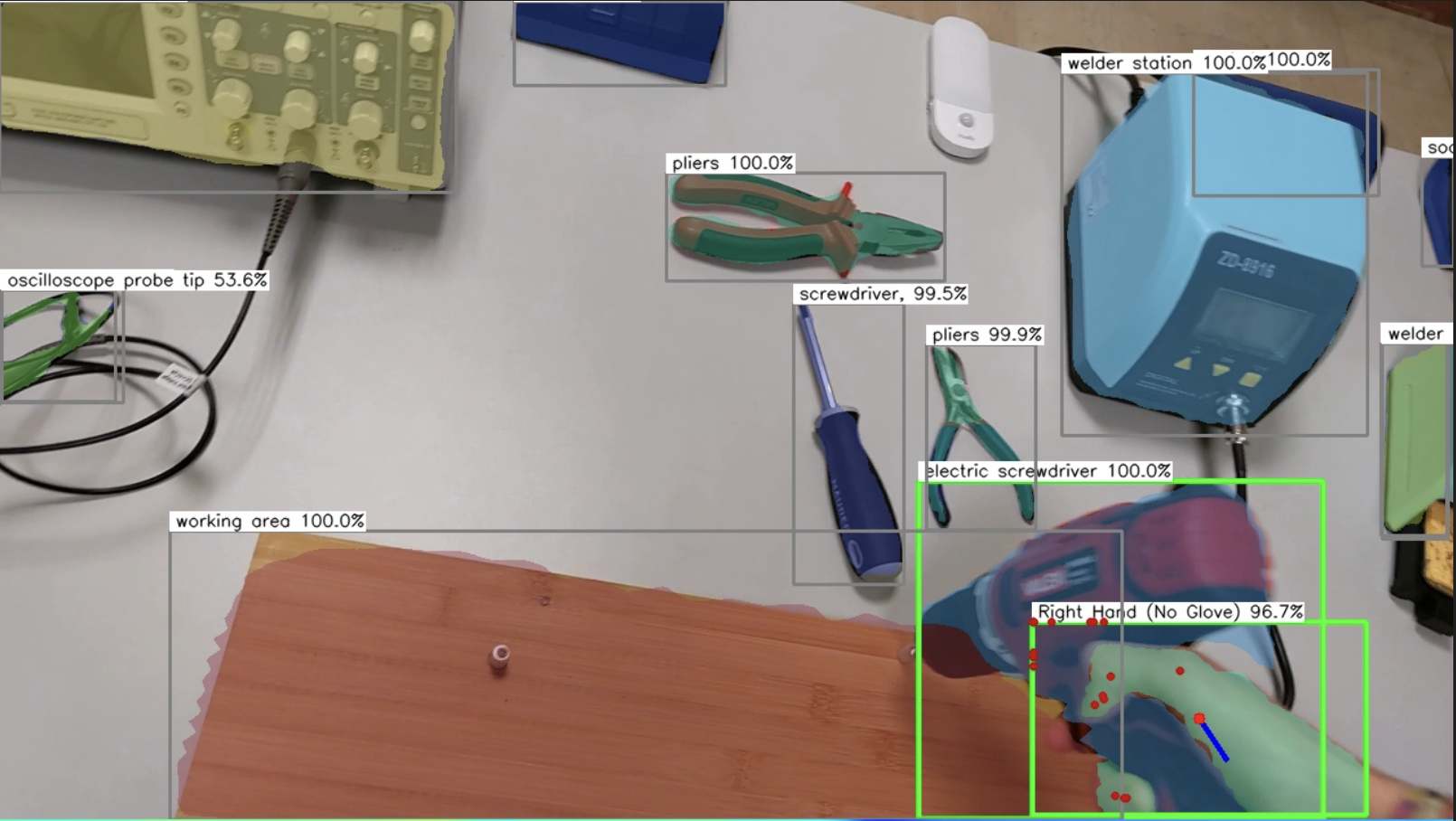}
    }
    \caption{Qualitative comparison between \textbf{GlovEgo-Net w/o Keypoints} (a) and the full \textbf{GlovEgo-Net} model (b). }
    \label{fig:zeroshot_ablation}

\end{figure*}

Qualitative comparisons across \textbf{GlovEgo-HOI-Synth} and \textbf{GlovEgo-HOI-Real} (Fig.~\ref{fig:qualitative_overview}) confirm that the \textbf{Synth+Real} model provides the most accurate predictions. In the synthetic domain, the \textbf{Real-Only} model struggles with object localization due to the domain gap, while \textbf{Synth-Only} and \textbf{Synth+Real} produce precise hand-object associations. 

In real-world scenarios, while all regimes achieve correct hand detection, both \textbf{Real-Only} and \textbf{Synth-Only} fail to accurately identify the manipulated object. Conversely, the \textbf{Synth+Real} model consistently correctly identifies the target object, confirming that \textbf{synthetic pre-training} combined with \textbf{real-world fine-tuning} yields superior performance across both domains.

Moreover, we provide additional qualitative examples to show the effect of the architectural design. As shown in Figure~\ref{fig:zeroshot_ablation}, the full \textit{GlovEgo-Net} correctly predicts both the hand contact state and the manipulated object, whereas the variant without the Keypoint Head fails to accurately associate the contact state with the correct spatial projection of the object, often resulting in interaction misses in complex scenes.
\FloatBarrier
\section{Conclusion}
\label{sec:conclusion}



We introduced \textit{GlovEgo-HOI}, a new benchmark for HOI detection in industrial environments, built using a synthetic data generation pipeline and a diffusion-based glove-augmentation process on real-world images. The dataset comprises two subsets, \textit{GlovEgo-HOI-Synth} and \textit{GlovEgo-HOI-Real}, providing rich multimodal annotations for hands and objects.

Alongside the dataset, we proposed \textit{GlovEgo-Net}, an architecture specifically designed for industrial EHOI detection. The model includes a \textit{Glove-Head} for recognizing PPE and a \textit{Keypoint-Head} for leveraging fine-grained hand keypoint information.

Extensive experiments on both the synthetic and real splits of our \textit{GlovEgo-HOI} benchmark demonstrate the effectiveness of the proposed framework. 

To foster further research and reproducibility, we publicly release the \textit{GlovEgo-HOI} dataset, our data generation and augmentation pipeline, and pretrained \textit{GlovEgo-Net} models at: {\iconlink{https://github.com/NextVisionLab/GlovEgo-HOI}{GitHub project.}}.
\section*{\uppercase{Acknowledgements}}

This research has been supported by Next Vision s.r.l. and by the project Future Artificial Intelligence Research (FAIR) - PNRR MUR Cod. PE0000013 - CUP: E63C22001940006.

\bibliographystyle{apalike}
{\small \bibliography{bibliography}}

\end{document}